\documentclass[lettersize,journal]{IEEEtran}

\usepackage{amsmath,amsfonts}
\usepackage{algorithmic}
\usepackage{algorithm}
\usepackage{array}
\usepackage[caption=false,font=normalsize,labelfont=sf,textfont=sf]{subfig}
\usepackage{textcomp}
\usepackage{stfloats}
\usepackage{url}
\usepackage{verbatim}
\usepackage{graphicx}
\usepackage{cite}

\usepackage{xcolor}

\usepackage{adjustbox}
\usepackage{booktabs}
\usepackage{multirow}
\usepackage{makecell}
\usepackage[colorlinks,
            linkcolor=blue,
            anchorcolor=blue,
            citecolor=blue,
            urlcolor=blue]{hyperref}

\begin{document}
\bstctlcite{setting}

\title{EndoMatcher: Generalizable Endoscopic Image Matcher via Multi-Domain Pre-training for Robot-Assisted Surgery}

\author{Bingyu Yang, Qingyao Tian, Yimeng Geng, Huai Liao, Xinyan Huang, Jiebo Luo
\IEEEmembership{Fellow, IEEE}, and Hongbin Liu
\thanks{Bingyu Yang, Qingyao Tian, and Yimeng Geng are with State Key Laboratory of Multimodal Artificial Intelligence Systems, Institute of Automation, Chinese Academy of Sciences, Beijing 100190, China, and also with the School of Artificial Intelligence, University of Chinese Academy of Sciences, Beijing 100049, China (e-mail: yangbingyu2022@ia.ac.cn; tianqingyao2021@ia.ac.cn; gengyimeng2022@ia.ac.cn).}
\thanks{Huai Liao, M.D. and Xinyan Huang, M.D. are with the Department of Pulmonary and Critical Care Medicine, The First Affiliated Hospital of Sun Yat-sen University, Guangzhou, Guangdong 510275, China (e-mail: liaohuai@mail.sysu.edu.cn; hxinyan@mail.sysu.edu.cn).}
\thanks{Jiebo Luo is with the Hong Kong Institute of Science \& Innovation, Hong Kong SAR during the sabbatical leave (e-mail: jluo@hkisi.org.hk).}
\thanks{Corresponding author: Hongbin Liu is with the Institute of Automation, Chinese Academy of Sciences, Beijing 100190, China, and also with the Centre of AI and Robotics, Hong Kong Institute of Science \& Innovation, Hong Kong SAR (e-mail: liuhongbin@ia.ac.cn).}
}

\markboth{Journal of \LaTeX\ Class Files,~Vol.~14, No.~8, August~2021}%
{Shell \MakeLowercase{\textit{et al.}}: A Sample Article Using IEEEtran.cls for IEEE Journals}

\IEEEpubid{0000--0000/00\$00.00~\copyright~2021 IEEE}
\maketitle

\begin{abstract}

Generalizable dense feature matching in endoscopic images is crucial for robot-assisted tasks, including 3D reconstruction, navigation, and surgical scene understanding. Yet, it remains a challenge due to difficult visual conditions (e.g., weak textures, large viewpoint variations) and a scarcity of annotated data. To address these challenges, we propose EndoMatcher, a generalizable endoscopic image matcher via large-scale, multi-domain data pre-training. To address difficult visual conditions, EndoMatcher employs a two-branch Vision Transformer to extract multi-scale features, enhanced by dual interaction blocks for robust correspondence learning. To overcome data scarcity and improve domain diversity, we construct Endo-Mix6, the first multi-domain dataset for endoscopic matching. Endo-Mix6 consists of approximately 1.2M real and synthetic image pairs across six domains, with correspondence labels generated using Structure-from-Motion and simulated transformations. The diversity and scale of Endo-Mix6 introduce new challenges in training stability due to significant variations in dataset sizes, distribution shifts, and error imbalance. To address them, a progressive multi-objective training strategy is employed to promote balanced learning and improve representation quality across domains. This enables EndoMatcher to generalize across unseen organs and imaging conditions in a zero-shot fashion. Extensive zero-shot matching experiments demonstrate that EndoMatcher increases the number of inlier matches by 140.69\% and 201.43\% on the Hamlyn and Bladder datasets over state-of-the-art methods, respectively, and improves the Matching Direction Prediction Accuracy (MDPA) by 9.40\% on the Gastro-Matching dataset, achieving dense and accurate matching under challenging endoscopic conditions. The code is publicly available at \url{https://github.com/Beryl2000/EndoMatcher}.

\end{abstract}


\section{Introduction}
\IEEEPARstart{W}{ith} rapid advances in medical imaging, endoscopy has become widely used for diagnosing and treating various diseases~\cite{nezhat2011nezhat}, as shown in Fig.~\ref{intro}(a). However, due to the narrow field of view and lack of depth, it remains difficult for clinicians to accurately localize and perceive anatomical targets. In recent years, robot-assisted systems have emerged as a promising solution, offering enhanced visual perception and operational precision, and are seen as key to improving surgical safety and intelligence~\cite{burgner2015continuum,morlana2024colonmapper,chen2024design}. Fig.~\ref{intro}(b) illustrates a typical setup of a monocular endoscope, along with sample image frames captured along a representative trajectory, commonly used as visual input in robotic platforms.

\IEEEpubidadjcol

Against this backdrop, image feature matching is essential for vision-based tasks. By establishing correspondences between image pairs, it enables camera pose estimation (R, T), which supports downstream applications such as image stitching~\cite{sharma2023comparative,lu2024s2p}, visual localization~\cite{roessle2023end2end,sarlin2020superglue,tian2024bronchotrack}, and 3D reconstruction~\cite{cadena2017past,morlana2024colonmapper,jiang2024few,schonberger2016structure}, as shown in Fig.~\ref{intro}(c). Earlier hand-crafted descriptors such as SIFT~\cite{lowe2004distinctive} and ORB~\cite{rublee2011orb} have proven effective in natural scenes. Recently, deep learning-based dense matching methods~\cite{detone2018superpoint,potje2023enhancing,revaud2019r2d2,lindenberger2023lightglue,sarlin2020superglue,sun2021loftr,truong2023pdc,jiang2024omniglue,shen2024gim} have significantly improved matching accuracy and density through end-to-end training, demonstrating stronger feature representation capabilities. However, when applied to endoscopic scenarios, both traditional and learning-based approaches suffer significant performance degradation and exhibit poor generalization due to two key challenges:

\begin{figure}[!t]
\centerline{\includegraphics[width=\columnwidth]{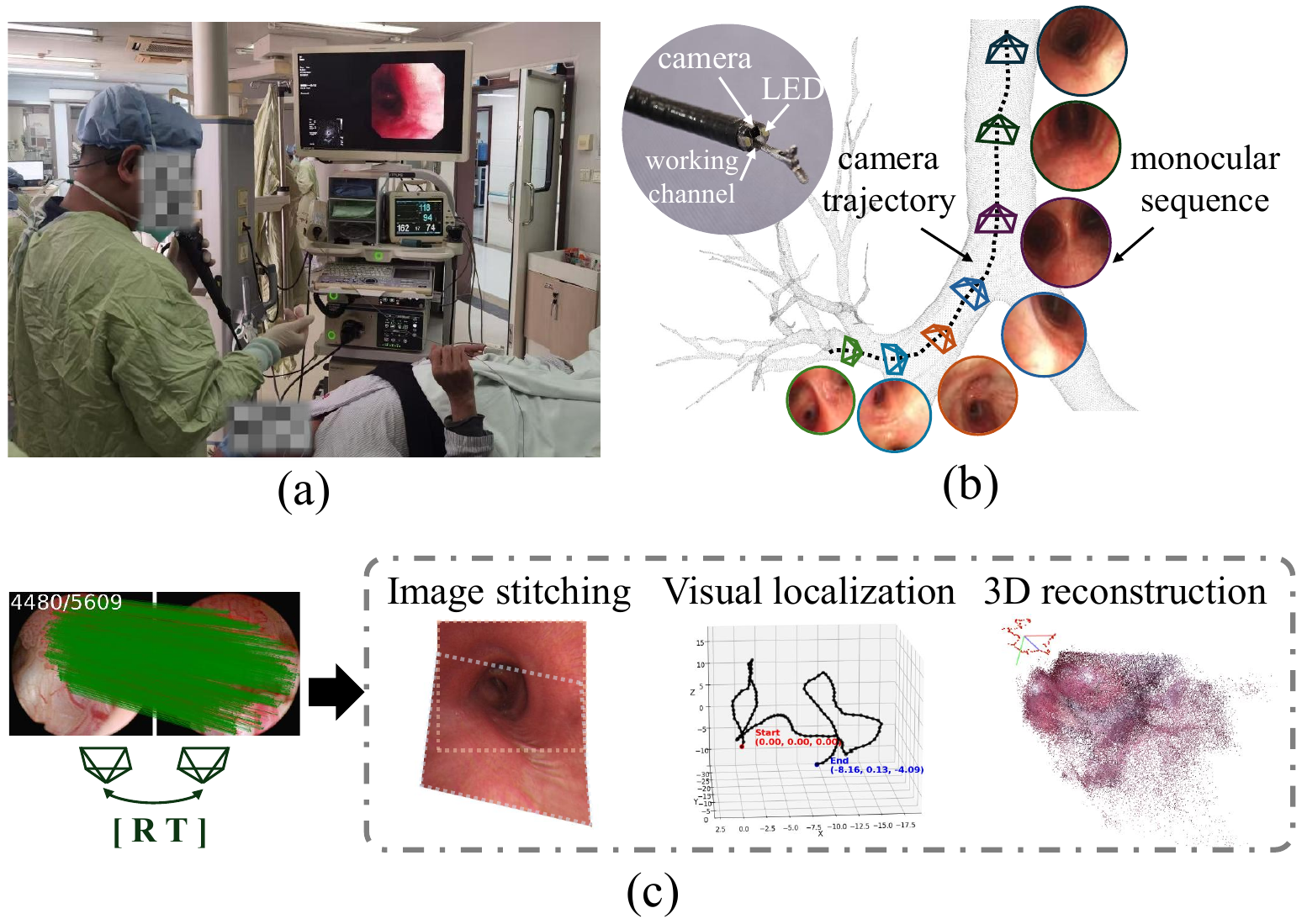}}
\caption{Endoscopic image matching in robot-assisted surgery. (a) Endoscopy used in clinical diagnosis for a patient. (b) Image sequence captured by a monocular endoscope along a representative trajectory. (c) Endoscopic image feature matching supporting robot-assisted visual tasks, including image stitching, visual localization, and 3D reconstruction. [R, T] denote camera pose parameters in rotation and translation.}
\label{intro}
\end{figure}

\noindent \textbf{Difficult visual conditions.} The mucosal surfaces observed in endoscopic images are often smooth or repetitive in texture. Rapid camera viewpoint and advancement within narrow anatomical passages can induce significant rotations exceeding 70°~\cite{lu2024s2p}. Under such conditions, traditional descriptors and shallow features struggle to extract stable keypoints, leading to unreliable correspondences or even complete failure in matching.

\noindent \textbf{Scarcity of annotated data.} Most deep matching models are trained on large-scale natural image datasets~\cite{sun2021loftr,jiang2024omniglue,shen2024gim}. However, acquiring ground-truth correspondences in endoscopy via RGB-D scanning~\cite{dai2017scannet} is limited by practical constraints, while manual annotation is time-consuming and error-prone~\cite{lu2024s2p}. 
Moreover, existing datasets lack diversity across organs, patients, and lighting, resulting in poor generalization and frequent failures in unseen scenarios.

To address these challenges, we propose EndoMatcher, a generalizable endoscopic image matcher pretrained on a large-scale and multi-domain dataset, Endo-Mix6. EndoMatcher enables dense and accurate matching under weak textures and significant viewpoint variations, while exhibiting strong zero-shot generalization across datasets.

Inspired by DPT ~\cite{ranftl2021vision}, which combines a ViT encoder with a CNN decoder to retain both global context and spatial precision for dense prediction, EndoMatcher employs a two-branch Siamese architecture based on the Vision Transformer to robustly learn correspondences under challenging visual conditions. Specifically, we reformulate descriptor learning as a keypoint localization task~\cite{liao2019multiview,liu2020extremely}. The hybrid-attention encoder is enhanced with dual interaction blocks to improve feature communication across image pairs. A pyramid-fusion decoder builds coarse-to-fine dense descriptors, while a multi-scale matching module generates pyramid response heatmaps according to source point features and matches points based on high responses. Training is supervised via a robust multi-scale response loss, which suppresses incorrect activations across scales and improves reliability under viewpoint variation.

To address data scarcity and improve domain diversity, we draw inspiration from foundation models in computer vision~\cite{ranftl2020towards,kirillov2023segment,yang2024depthv2}. MiDas~\cite{ranftl2020towards} highlights that diverse training data is key to generalization. Depth Anything v2~\cite{yang2024depthv2} further confirms that high-quality synthetic labels benefit detailed predictions. Motivated by these, we construct Endo-Mix6, the first large-scale multi-domain dataset for endoscopic feature matching, containing approximately 1.2M image pairs from six real and synthetic sources. It covers diverse organs, viewpoint variations, and lighting conditions, serving as a pre-training benchmark for zero-shot generalization. Given that manual annotation is time-consuming and error-prone, we generate labels using Structure-from-Motion (SfM)~\cite{schonberger2016structure}. Although widely used in natural scenes~\cite{sun2021loftr, lindenberger2023lightglue,jiang2024omniglue, shen2024gim,zhu2025transmatch}, SfM-generated labels are often sparse and noisy on endoscopic data, reducing their reliability and limiting training performance. To address this, we further simulate camera viewpoints using perspective transformations for data augmentation.

Significant discrepancies in dataset sizes and imbalanced-error within Endo-Mix6 pose new challenges to training stability. To maximize the benefits of hybrid training, we explore a progressive multi-objective training strategy. EndoMatcher is first trained on synthetic data for stable feature learning, then fine-tuned with real data to enhance realism and diversity. This strategy balances domain contributions and boosts generalization to unseen organs and clinical scenarios.

In summary, our main contributions are as follows:
\begin{itemize}
\item We propose EndoMatcher, a generalizable endoscopic image matcher that employs a two-branch Vision Transformer enhanced with dual interaction blocks to extract multi-scale features, enabling dense and accurate matching under challenging conditions such as weak textures and large viewpoint variations.
\item We introduce Endo-Mix6, the first large-scale multi-domain dataset for endoscopic feature matching, comprising approximately 1.2M real and synthetic image pairs across six domains. Labels are automatically generated via SfM and simulated transformations, requiring no manual annotation. A progressive multi-objective training strategy is designed to encourage balanced learning and improve zero-shot generalization.
\item Experimental results show that EndoMatcher significantly outperforms existing SOTA methods in challenging regions in endoscopic images, demonstrating strong performance in dense matching and excellent transferability in zero-shot endoscopic feature matching tasks.
\end{itemize}

\section{RELATED WORK}

\subsection{Image Matching Methods}
Traditional image matching algorithms relied on hand-crafted local features~\cite{lowe2004distinctive,rublee2011orb,liu2022improved,xie2020endoscope}. These methods typically involve keypoint detection, descriptor extraction, or nearest-neighbor matching. While generally robust across diverse scenarios, they struggle with weak textures or repetitive structures. Deep learning has revolutionized the field of image matching~\cite{detone2018superpoint,revaud2019r2d2,truong2023pdc,potje2023enhancing,sarlin2020superglue,sun2021loftr,lindenberger2023lightglue, bokman2022case, chen2022aspanformer,lu2024s2p,jiang2024omniglue,zhu2025transmatch}. SuperGlue~\cite{sarlin2020superglue} enhances keypoint descriptors with SuperPoint~\cite{detone2018superpoint} and matches sparse features using attention, but its performance is limited by keypoint detection quality~\cite{sun2021loftr}. Detector-free methods like LoFTR~\cite{sun2021loftr} overcome this by directly estimating dense correspondences via Transformer-based representations.

However, most existing models are trained on natural images and lack optimization for medical scenarios. To address the challenges of endoscopic image matching, Liu et al.~\cite{liu2020extremely} proposed a self-supervised dense descriptor learning framework that improves matching performance in sinus endoscopy by incorporating globally aware descriptors. Lu et al.~\cite{lu2024s2p} introduced S2P-Match for magnetically controlled capsule endoscopy (MCCE), combining contrastive learning and Transformers to perform patch-level matching. However, it lacks pixel-level precision and shows limited generalization beyond MCCE-specific scenarios. TransMatch~\cite{zhu2025transmatch} addresses severe deformations in gastroscopy using a Transformer combined with a bridging strategy and a dedicated deblurring module to enhance matching robustness. Despite its effectiveness, the approach incurs high inference cost and is tailored to specific task conditions. In contrast, our method adopts a detector-free Transformer via multi-domain pre-training, enabling robust and dense matching across diverse endoscopic data.

\begin{figure*}[!t]
\centerline{\includegraphics[width=\textwidth]{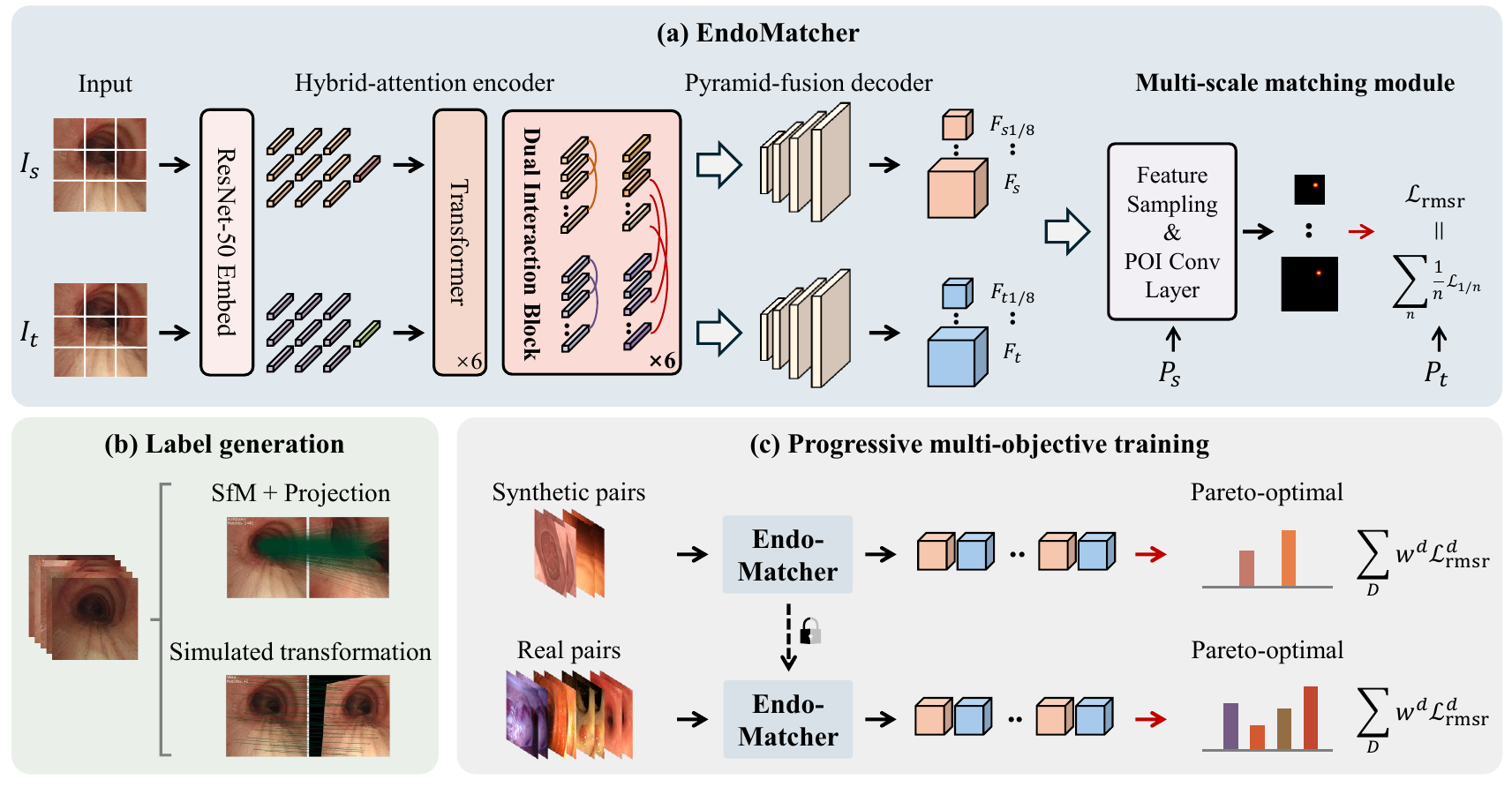}}
\caption{Overview of the proposed method. (a) Network architecture of EndoMatcher, which includes a hybrid-attention encoder with dual interaction blocks (DIB), a pyramid-fusion decoder, and a multi-scale matching module designed to handle large viewpoint variations. (b) Label generation pipeline of Endo-Mix6, combining structure-from-motion (SfM) reconstruction and simulated transformations. (c) A progressive multi-objective training strategy that promotes domain generalization across diverse endoscopic data.}
\label{fig:overview}
\end{figure*}

\subsection{Zero-shot Learning with Large Vision Models}

Large vision models trained on diverse datasets exhibit strong cross-domain generalization~\cite{ranftl2020towards,kirillov2023segment,oquab2023dinov2, birkl2023midas, jiang2024omniglue,shen2024gim, yang2024depth, tian2024endoomni, tian2025endomamba}. 
SAM~\cite{kirillov2023segment}, trained on a vast corpus for segmentation, demonstrated impressive out-of-distribution performance. DINOv2~\cite{oquab2023dinov2} employs self-supervised pre-training on curated multi-source image data, yielding general-purpose visual features with strong multi-task generalization. For image matching tasks, large-scale vision models are increasingly adopted to improve generalization. OmniGlue~\cite{jiang2024omniglue} integrates foundation models into natural image matching pipelines, achieving a ~20\% accuracy boost in cross-domain scenarios. 
% GIM~\cite{shen2024gim} is trained on large-scale internet video data, enabling cross-domain matching without fine-tuning. 

While foundation models trained on large-scale, multi-source data show strong cross-domain generalization, their effectiveness often depends on how the training data is mixed. Naive combination of diverse datasets can cause domain bias and conflicting objectives, limiting generalization. In monocular depth estimation, MiDaS~\cite{ranftl2020towards} adopt scale-invariant losses and multi-objective learning to align cross-domain features, while Metric3D v2~\cite{hu2024metric3d} employs canonical transformations and joint depth-normal optimization to handle diverse camera settings. However, large-scale diverse datasets for endoscopic matching are rare, and systematic studies on cross dataset learning remain limited.

To address these limitations, we present EndoMatcher, a generalizable endoscopic image matcher based on a two-branch Vision Transformer enhanced with dual interaction blocks. We construct Endo-Mix6, a large-scale, multi-domain dataset with real and synthetic image pairs. EndoMatcher is pretrained using a progressive multi-objective training strategy. Extensive experiments show that EndoMatcher significantly improves generalization, particularly under weak textures and large viewpoint changes.

\section{METHOD}

In this study, we propose EndoMatcher, a novel and generalizable endoscopic image matcher designed to address the challenges of feature matching in diverse endoscopic scenarios, including weak textures, large viewpoint variations, and limited annotated data. As illustrated in Fig.~\ref{fig:overview}, EndoMatcher overcomes visual condition challenges with three key network components: a hybrid-attention encoder, a pyramid-fusion decoder, and a multi-scale matching module. To enable zero-shot generalization across diverse datasets, EndoMatcher is pretrained using a progressive multi-objective training strategy on Endo-Mix6, a large-scale and multi-domain dataset. Endo-Mix6 comprises six domains from both synthetic and real data sources, with labels generated via SfM reconstruction and simulated transformations.

\subsection{EndoMatcher}
We design EndoMatcher, a ViT-based dual-branch Siamese architecture to process a pair of endoscopic images, $I_s$, $I_t$ (source and target images, respectively), enabling robust correspondence learning under challenging visual conditions, as shown in Fig.~\ref{fig:overview}(a). Specifically, we convert descriptor learning as a keypoint localization task~\cite{liao2019multiview,liu2020extremely}. To enhance cross-image feature interaction, we incorporate Dual Interaction Blocks (DIB) into the hybrid-attention encoder. The pyramid-fusion decoder then generates coarse-to-fine descriptors, which are supervised by the multi-scale matching module. This module computes target pyramid response heatmaps from source keypoint $P_s$, and a robust multi-scale response loss (RMSR) is introduced to suppress false activations of the target location $P_t$, enabling reliable matching under significant viewpoint changes.

\begin{figure}[!t]
\centerline{\includegraphics[width=\columnwidth]{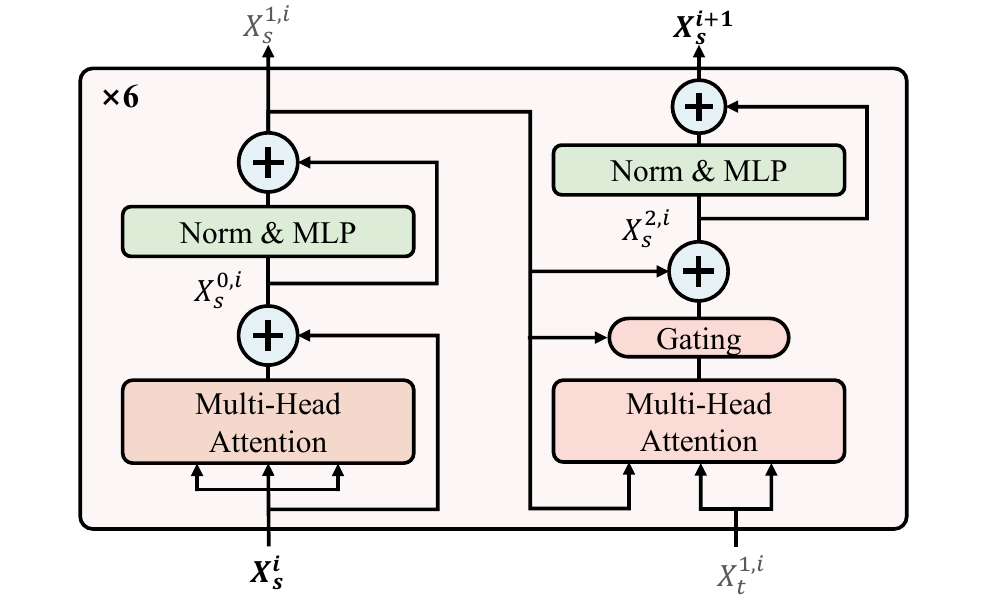}}
\caption{Architecture of the Dual Interaction Block (DIB). Each block contains the intra-branch self-attention and the inter-branch gated cross-attention. Cross-attention is modulated by a learnable gate to suppress noisy or irrelevant inter-image information. $X^i_s$ and $X^{i+1}_s$ represent the input and output of the $i$-th DIB, respectively. The intermediate result $X^{1,i}_{s(t)}$ is used for dual-branch interaction.}
\label{dib}
\end{figure}

\subsubsection{Hybrid-attention encoder}

Each branch employs a shared ResNet-50 backbone~\cite{he2016deep} to extract pixel-level features at 1/16 of the input resolution, which are then converted into tokens and fed into the Vision Transformer~\cite{ranftl2021vision}. Unlike the original ViT with 12 standard transformer blocks, our Hybrid-attention Encoder consists of two stages: the first 6 blocks use self-attention only, while the final 6 are replaced by dual interaction blocks (DIBs), which alternate self-attention and cross-attention to enhance early feature interaction between the two image branches. Specifically, each block consists of an intra-branch self-attention module and an inter-branch gated cross-attention module, followed by a feed-forward network with residual connections. Given source and target tokens $X^i_s, X^i_t \in \mathbb{R}^{N \times D}$ at the $i$-th DIB, the update rule for the source token $X^i_s$ is defined as:
\begin{equation}
X^{0,i}_s = \mathrm{MHA}(X^i_s) + X^i_s,
\end{equation}
\begin{equation}
X^{1,i}_s = \mathrm{MLP}(\mathrm{LN}(X^{0,i}_s) + X^{0,i}_s,
\end{equation}
\begin{equation}
X^{2,i}_s = X^{1,i}_s + G \odot \mathrm{MHA}(X^{1,i}_s, X^{1,i}_t, X^{1,i}_t) ,
\end{equation}
\begin{equation}
X^{i+1}_s = \mathrm{MLP}(\mathrm{LN}(X^{2,i}_s)) + X^{2,i}_s,
\end{equation}
where $\mathrm{MHA}(\cdot)$ denotes multi-head attention, $\mathrm{MLP}(\cdot)$ is a two-layer feed-forward network, and $\mathrm{LN}(\cdot)$ is layer normalization~\cite{dosovitskiy2020image}. $\odot$ denotes element-wise multiplication. $X^{0,i}_s$, $X^{1,i}_s$, and $X^{2,i}_s$ are intermediate results. To enable more reliable cross-view information exchange, each DIB includes a gated cross-attention module that acts as a content-aware filter, selectively integrating informative features while suppressing noisy or ambiguous signals. The gating map is computed as $G = \sigma\left(W_2 \cdot \mathrm{ReLU}(W_1 \cdot X^{1,i}_s)\right)$, where $W_1$ and $W_2$ are learnable projections and $\sigma(\cdot)$ is the sigmoid activation. The same operation is applied symmetrically to the target branch.

% Each branch employs a shared ResNet-50 backbone~\cite{he2016deep} to extract pixel-level features at 1/16 of the input resolution, which are then converted into tokens and fed into the Vision Transformer~\cite{ranftl2021vision}. Unlike the original ViT with 12 standard transformer blocks, our Hybrid-attention Encoder consists of two stages: the first 6 blocks use self-attention only, while the final 6 are replaced by dual interaction blocks (DIB) that alternate self-attention and cross-attention to enhance early feature interaction between the image pair. This design effectively models long-range dependencies and contextual interactions, which is especially beneficial for endoscopic scenes with weak textures and repetitive structures.

\begin{table*}[t!]
\centering
\caption{Overview of the datasets used in our work. Top: Endo-Mix6. Bottom: Zero-shot test sets.}
\label{tab:datasets}
\setlength{\tabcolsep}{22pt}  
\renewcommand{\arraystretch}{1.5} 
\begin{tabular}{ccccc}
\toprule
\textbf{Dataset} & \textbf{Organs} & \textbf{Seqs}  & \textbf{Frames} & \textbf{Pairs (r=20)} \\
\midrule
C3VD~\cite{bobrow2023colonoscopy} & Colon (phantom) & 22 & 3,218 & 59,740 \\
EndoSLAM~\cite{ozyoruk2021endoslam} & Colon, stomach, etc. (simulation) & 340 & 34,306 & 614,720 \\
SCARED~\cite{allan2021stereo} & Abdomen & 108 & 13,712 & 251,560 \\
EndoMapper~\cite{azagra2023endomapper} & Colon & 27 & 3,406 & 62,450 \\
Colonoscopic~\cite{mesejo2016computer} & Colon & 21 & 2,724 & 50,070 \\
Ours-Bronch & Airway & 68 & 8233 & 150,380 \\
\midrule 
Hamlyn Centre Dataset$^\dagger$ & Kidney & 55 & 7,532 & 139,090 \\
Bladder Tissue Dataset~\cite{lazo2023semi} & Bladder & 32 & 539 & 8,600 \\
Gastro-matching Dataset~\cite{zhu2025transmatch} & Stomach & 190 & 1,271 & 12,140 \\
\bottomrule
\end{tabular}

% \vspace{0.3em}
\begin{flushleft}
\scriptsize\textbf{$^\dagger$} \url{http://hamlyn.doc.ic.ac.uk/vision/}.
\end{flushleft}

\end{table*}

\begin{figure*}[t!]
\centerline{\includegraphics[width=\textwidth]{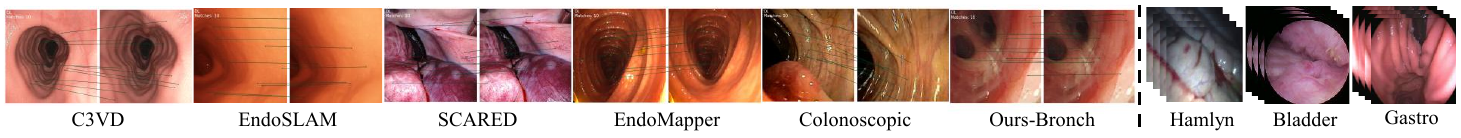}}
\caption{Sample point correspondences from the six pre-training datasets used to supervise the model, along with sample frames from the three testing datasets employed for zero-shot matching.}
\label{fig:dataset}
\end{figure*}

\subsubsection{Pyramid-fusion decoder}

In the pyramid-fusion decoder, EndoMatcher extracts multi-scale information from both early ResNet layers (the first and second blocks) and Transformer tokens (the 3th and 6th DIB). These tokens are reassembled into image-like feature maps via the Reassemble modules~\cite{ranftl2021vision}. The resulting features are progressively upsampled and refined using RefineNet-based fusion blocks~\cite{lin2017refinenet} with residual learning, preserving spatial details throughout. This process yields pyramid feature maps at 1/8, 1/4, and 1/2 resolutions, as well as a final full-resolution feature map produced by an additional upsampling head.

\subsubsection{Multi-scale matching module}

To ensure robust matching under large viewpoint variations, a multi-scale matching module is employed in EndoMatcher, operating on pyramid feature maps at resolutions of $1/8$, $1/4$, $1/2$, and full scale. Specifically, given multi-scale descriptors $F_{s1/n}$ and $F_{t1/n}$, a source keypoint descriptor $F_{s1/n}(P_s)$ is sampled at location $P_s$. This descriptor is treated as a $1 \times 1$ convolutional kernel to perform a Point-of-Interest (POI) convolution~\cite{liao2019multiview} over the target feature map $F_{t1/n}$, producing a heatmap $M_{t1/n}$ that encodes the similarity between $F_{s1/n}(P_s)$ and all target descriptors.

To supervise the learning of reliable correspondences across scales, a robust multi-scale response (RMSR) loss is introduced. This loss encourages strong activation at the true correspondence location $P_t$, while suppressing responses at other positions in the predicted heatmap $M_{t1/n}$ at each scale. It is formulated as:
\begin{equation}
\mathcal{L}_{\mathrm{rmsr}} = \sum_{n} \frac{1}{n} \left( -\log\left( \frac{e^{\sigma M_{t1/n}(P_t)}}{\sum_{P} e^{\sigma M_{t1/n}(P)}} \right) \right),
\label{eq5}
\end{equation}
where $\sigma$ is a temperature scaling factor and $n \in \{1, 2, 4, 8\}$ denotes the resolution level. This multi-scale supervision provides richer contextual cues, effectively improving robustness against geometric and appearance variations.

Furthermore, to improve training robustness, a residual clipping strategy is employed to discard the lowest 20\% of response errors, focusing optimization on harder samples and thus enhancing the model’s resilience to outliers and challenging conditions.

\subsubsection{Dense feature matching}
During inference, EndoMatcher can directly handle any pair of endoscopic images without requiring domain-specific tuning. The model extracts deep features to generate a full-resolution target heatmap for each keypoint in the source image, where the position with the highest response in the heatmap is taken as the estimated location of the corresponding target keypoint. To further improve matching accuracy, EndoMatcher incorporates a cycle consistency principle~\cite{liu2020extremely}, which verifies the mutual correspondence of matched points in both directions, effectively suppressing errors and outliers caused by one-way matching. For downstream tasks that require high geometric consistency, such as 3D reconstruction, the system can optionally apply RANSAC for geometric verification, eliminating matched point pairs that violate geometric constraints from the perspective of global spatial structure.

\subsection{Cross Dataset Learning}

To address the scarcity of annotations and enhance generalization across diverse endoscopic matching tasks, we construct Endo-Mix6, a large-scale, multi-domain dataset. To overcome training stability due to significant variations in dataset sizes, we propose a progressive multi-objective training strategy to promote balanced learning.

\subsubsection{Endo-Mix6}

As summarized in Table~\ref{tab:datasets}, Endo-Mix6 integrates data from six public and clinical sources, including a newly collected bronchoscopic subset from our collaborating hospital. Raw videos are segmented into 5-second clips at 30 fps, and image pairs are sampled with a maximum temporal interval of 20 frames (r=20). Endo-Mix6 comprises approximately 1.2M image pairs spanning various domains, including gastroscopy, colonoscopy, laparoscopy, and bronchoscopy. The Hamlyn Centre, Bladder Tissue, and Gastro-matching datasets are excluded from training and reserved exclusively for zero-shot evaluation on previously unseen anatomical regions. Among them, the Hamlyn Centre and Bladder Tissue datasets are unlabeled, while Gastro-Matching includes 150 image pairs with 60 ground-truth correspondences and 40 video clips with camera movement directions. Examples of point correspondences and test images are shown in Fig.~\ref{fig:dataset}.

For synthetic datasets like EndoSLAM and C3VD, dense pixel-level correspondences are obtained from ground-truth camera poses and depth maps. To avoid the cost and noise of manual annotations on unlabeled clinical data, we adopt two automatic label-generation strategies: Structure-from-Motion (SfM)~\cite{schonberger2016structure} and simulated perspective transformations, as illustrated in Fig.~\ref{fig:overview}(b). The first uses a LoFTR-enhanced SfM pipeline to reconstruct 3D point clouds from short video sequences. These points are then back-projected to 2D frames, with visibility checks applied to retain valid projections. To improve label density, we include visibility from neighboring frames and apply RANSAC filtering with a strict reprojection threshold to ensure geometric consistency. The second strategy creates dense synthetic correspondences by applying random perspective transformations. Corner perturbations and simulated translations are used to generate realistic viewpoint shifts, followed by pixel sampling and warping to obtain point correspondences. Correspondences falling outside the valid region or suffering from excessive distortion are discarded.

\begin{table*}[!tb]
\caption{Zero-shot dense correspondence quality and inference speed (FPS) on two endoscopic datasets. FPS for the last three methods is measured per 1K matches. The \textbf{best} and \underline{second-best} are highlighted.}
\label{tab:comparison}
\centering
\renewcommand{\arraystretch}{1.5} 
\setlength{\tabcolsep}{6pt}       
\begin{tabular}{c|c|ccc|ccc|c}
\toprule
\multirow{2}{*}{} & \multirow{2}{*}{Training Sets}
& \multicolumn{3}{c|}{\textbf{Hamlyn Centre Dataset}} 
& \multicolumn{3}{c|}{\textbf{Bladder Tissue Dataset}} 
& \multirow{2}{*}{FPS $\uparrow$} \\
\cline{3-8}
& & $N_{\text{pt}}$ & $N_{\text{inlier}}$ & $\text{KR}$ & $N_{\text{pt}}$ & $N_{\text{inlier}}$ & $\text{KR}$ & \\
\midrule
SIFT~\cite{lowe2004distinctive} + FLANN~\cite{muja2009flann} & - & 5.97 & 4.15 & 28.67 & 18.34 & 13.52 & 50.23 &-  \\
SuperPoint~\cite{detone2018superpoint} + SuperGlue~\cite{sarlin2020superglue} {\scalebox{0.8}{(CVPR20)}}  &COCO~\cite{lin2014microsoft} + ScanNet~\cite{dai2017scannet}  &47.80 &30.85  &60.74 &36.75  &26.31  &61.77  &6.11 \\
SuperPoint~\cite{detone2018superpoint} + LightGlue~\cite{lindenberger2023lightglue} {\scalebox{0.8}{(ICCV23)}}  &COCO~\cite{lin2014microsoft} + MegaDepth~\cite{li2018megadepth}  &230.39 &168.60  &69.45 &312.83  &223.37  &67.18  &8.91 \\
OmniGlue~\cite{jiang2024omniglue} {\scalebox{0.8}{(CVPR24)}} &MegaDepth~\cite{li2018megadepth}  &588.38 &225.34 &37.49 &616.57  &265.19  &41.74  &0.57 \\
LoFTR~\cite{sun2021loftr} {\scalebox{0.8}{(CVPR21)}} & ScanNet~\cite{dai2017scannet} &1847.44  &1486.84  &\underline{72.26}  & 1845.64 & 1382.48 & \underline{68.19} &\underline{12.83} \\
TransMatch~\cite{zhu2025transmatch} {\scalebox{0.8}{(TMI25)}} &MegaDepth~\cite{li2018megadepth}  &\underline{4676.61} &\underline{2648.92}  &55.25 &\underline{4557.93}  &\underline{2029.05}  &43.84  &0.02 \\
% GIM~\cite{shen2024gim} {\scalebox{0.8}{(ICLR24)}} & Internet Videos & \underline{4757.06} & \underline{3598.12} & \textbf{74.63} & \underline{4978.29} & \underline{3527.96} & \textbf{70.24} & \\
Ours & Endo-Mix6 &\textbf{8190.21} &\textbf{6375.70} 	&\textbf{73.40}  &\textbf{8346.82} &\textbf{6116.20} &\textbf{68.75}  &\textbf{47.38} \\

\bottomrule
\end{tabular}
\end{table*}

\begin{figure*}[t!]
\centerline{\includegraphics[width=\textwidth]{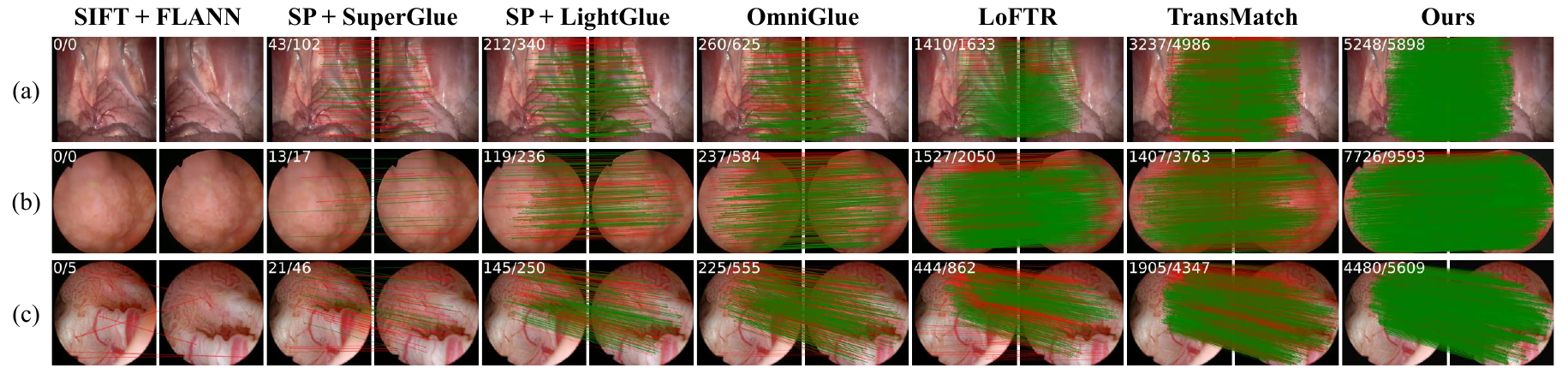}}
\caption{Visual comparisons of zero-shot feature matching results between our method and SOTA methods. From top to bottom: (a) image pairs with solid textures and large viewpoint shift; (b) consecutive images with weak textures and small viewpoint variation; (c) two images with weak textures and significant camera advancement. Inlier matches are shown in \textcolor[rgb]{0.1,0.5,0.32}{green}, while \textcolor{red}{red} lines indicate initial matches rejected by RANSAC. The white text in the top-left corner of each image displays $N_{\text{inlier}}/N_{\text{pt}}$.}
\label{fig:exper}
\end{figure*}

\subsubsection{Progressive multi-objective training}

To effectively bridge the domain gap between synthetic and real-world endoscopic data, we propose a progressive multi-objective training strategy designed to accelerate model convergence and enhance generalization, as illustrated in Fig.~\ref{fig:overview}(c). In the first stage, the model is pretrained on high-fidelity synthetic datasets, which provide abundant and noise-free supervision, enabling the network to rapidly learn stable and transferable representations. In the second stage, the pretrained model is fine-tuned on diverse real clinical datasets to adapt to domain-specific characteristics such as organ textures, lighting variations, and imaging noise. 

In both stages, each dataset is treated as a distinct task, and a shared set of parameters $\boldsymbol{\theta}$ is optimized using a Pareto-optimal multi-objective method~\cite{sener2018multi,ranftl2020towards}. Rather than minimizing a weighted sum of all dataset-specific losses, this framework seeks a solution where no individual loss can be further reduced without increasing at least one of the others. The training objective is formulated as:
\begin{equation}
\min_{\boldsymbol{\theta}}\left(\mathcal{L}_{\mathrm{rmsr}}^1(\boldsymbol{\theta}),\ldots,\mathcal{L}_{\mathrm{rmsr}}^D(\boldsymbol{\theta})\right)^\top,
\label{eq6}
\end{equation}
where $D$ denotes the number of datasets involved in the current training phase.

To approximate the Pareto front, the Multiple Gradient Descent Algorithm (MGDA) is proposed to compute a task-wise weight $w^d$ at each training step based on the current parameter state $\boldsymbol{\theta}$~\cite{sener2018multi}. These weights implicitly balance the gradient contributions from each dataset by identifying a descent direction that lies on the Pareto front. Accordingly, EndoMatcher is supervised by the following total loss:
\begin{equation}
\mathcal{L} = \sum_{D} w^d \mathcal{L}_{\mathrm{rmsr}}^d(\boldsymbol{\theta}),
\label{eq7}
\end{equation}
where $w^d$ is the optimal weight assigned to the $d$-th dataset. 

By leveraging this optimization strategy, our approach promotes balanced training across domains, mitigates negative transfer, and prevents overfitting to dominant datasets.

\section{EXPERIMENT}

We conducted extensive experiments on unseen datasets to evaluate the model’s feature matching capability under zero-shot learning scenarios. Additionally, we systematically analyzed the effects of different loss functions and encoder architectures through ablation studies. Furthermore, we investigated the impact of multi-dataset mixed training strategies on model performance.

\subsection{Implementation Details}

We first pretrain the model on synthetic datasets with high-precision labels using the AdamW optimizer (learning rate set to 5.0e-6), a batch size of 16, and train for 10 epochs. The model is then fine-tuned on real datasets, during which the ResNet50 backbone in the encoder is frozen. Fine-tuning is performed for 30 epochs with a batch size of 16, using a cosine schedule with a 5-epoch warm-up. All input image pairs are resized to 384×384 pixels, and 10 correspondences are randomly sampled from each pair as supervision. Data augmentation includes random adjustments to brightness (±20\%) and contrast (±30\%), motion blur, Gaussian noise, and elastic deformation to simulate local non-rigid motion. The scale factor $\sigma$ in the loss function is set to 20. To improve evaluation accuracy and reliability, parts of the validation and test sets from real data are manually curated and annotated. 

All dense matching operations are GPU-accelerated to improve both training and inference efficiency. A cycle-consistency constraint is introduced, allowing matched positions to vary within a local neighborhood around each keypoint, thereby enhancing robustness. 3D reconstruction is performed using COLMAP with default settings unless otherwise specified. The model is implemented in PyTorch and trained on a workstation equipped with an NVIDIA A800 GPU. All inference experiments are conducted using a GPU with 24GB of memory.

\subsection{Evaluation Metrics}

To comprehensively evaluate the model’s performance in both keypoint localization and dense matching, we adopt the Percentage of Correct Keypoints (PCK)~\cite{liu2020extremely} to assess localization accuracy, and introduce the number of initial matching points ($N_{\text{pt}}$), the number of inlier matches ($N_{\text{inlier}}$), and the Keep Ratio (KR)~\cite{liu2020extremely} to assess the density of matching.

PCK measures the percentage of predicted keypoints that fall within a normalized distance threshold of the ground truth. In ablations on loss functions and encoder designs, we report PCK at 5, 10, and 20 pixels. The metrics $N_{\text{pt}}$, $N_{\text{inlier}}$, and KR do not require ground truth and are widely used to evaluate correspondence density in endoscopic matching~\cite{liu2020extremely}, making them suitable for zero-shot comparison experiments on unlabeled datasets such as the Hamlyn Centre and Bladder Tissue datasets. Specifically, KR measures the proportion of correspondences retained after geometric verification:
\begin{equation}
\text{KR} = \frac{N_{\text{inlier}}}{N_{\text{pt}}},
\label{eq:KR}
\end{equation}
where $N_{\text{inlier}}$ denotes the number of correspondences that pass geometric verification (e.g., using RANSAC with a reprojection error below 5 pixels), and $N_{\text{pt}}$ is the total number of initial matches between image pairs.

To assess matching accuracy in downstream tasks, we further adopt the Homography Estimation Accuracy (HEA) and Matching Direction Prediction Accuracy (MDPA)~\cite{zhu2025transmatch}, and evaluate them on the Gastro-Matching dataset with manually annotated correspondences and motion directions. HEA quantifies the proportion of pixels in the reference image that are correctly projected to the target image via the estimated homography, under a given pixel error threshold. We report HEA@3px and HEA@5px in our experiments. MDPA assesses whether the predicted correspondences provide sufficient geometric cues to infer the camera’s motion directions, thereby reflecting the accuracy of the matching results.

\begin{table}[t]
\centering
\caption{Zero-shot matching accuracy on the Gastro-Matching Dataset using downstream metrics: HEA (\%) at 3px and 5px thresholds and MDPA. The \textbf{best} and \underline{second-best} are highlighted.}
\label{tab:acc}
\setlength{\tabcolsep}{5.5pt}
\renewcommand{\arraystretch}{1.4}
\begin{tabular}{c|ccc}
\toprule
\multirow{2}{*}{} & \multicolumn{3}{c}{\textbf{Gastro-matching Dataset}} \\
\cline{2-4} 
& HEA@3px & HEA@5px & MDPA \\
\midrule
SIFT~\cite{lowe2004distinctive} + FLANN~\cite{muja2009flann} &42.0 &57.1 &1.1 \\
SP~\cite{detone2018superpoint} + SuperGlue~\cite{sarlin2020superglue} {\scalebox{0.8}{(CVPR20)}} &78.0 &88.1 &59.6 \\
SP~\cite{detone2018superpoint} + LightGlue~\cite{lindenberger2023lightglue} {\scalebox{0.8}{(ICCV23)}}  &75.3 &86.0 &57.6\\
OmniGlue~\cite{jiang2024omniglue} {\scalebox{0.8}{(CVPR24)}} &70.4 &82.5 &52.8 \\
LoFTR~\cite{sun2021loftr} {\scalebox{0.8}{(CVPR21)}}  &71.8 &83.9  &53.8  \\
TransMatch~\cite{zhu2025transmatch} {\scalebox{0.8}{(TMI25)}}  &\underline{85.1} &\underline{93.3} &\underline{76.0}  \\
Ours  &\textbf{88.7} &\textbf{95.8}  &\textbf{85.4} \\
\bottomrule
\end{tabular}
\end{table}

\subsection{Zero-shot Feature Matching} % change to: Zero-shot generalization ?

\subsubsection{Matching density and efficiency}

We compare our EndoMatcher with SOTA methods across natural and medical domains to assess its generalization to unseen clinical scenarios. As shown in Table~\ref{tab:comparison}, we evaluate  the matching density and inference efficiency of zero-shot feature matching on two endoscopic datasets: The Hamlyn Centre and Bladder Tissue datasets. SIFT~\cite{lowe2004distinctive} fail to obtain sufficient inliers on either dataset. Sparse methods such as SuperGlue~\cite{sarlin2020superglue}, LightGlue~\cite{lindenberger2023lightglue}, and OmniGlue~\cite{jiang2024omniglue} extract keypoints using SuperPoint~\cite{detone2018superpoint} and match at most 2048 keypoints, which inherently limits their match density and scene coverage in low-texture endoscopic environments. OmniGlue improves generalization via foundation model guidance and keypoint-aware attention, but its complexity leads to slower inference. As a dense matcher, LoFTR~\cite{sun2021loftr} achieves the second-highest keep ratio (KR) on both datasets, but it generalizes poorly to medical domains. TransMatch~\cite{zhu2025transmatch}, designed specifically for gastroscopy, attains the second-highest match density but suffers from low KR on unseen organs and extremely slow inference due to its heavy architecture, limiting its scalability to long sequences. In contrast, EndoMatcher achieves the best overall performance across all metrics. It improves $N_{\text{inlier}}$ over TransMatch by 140.69\% on the Hamlyn Centre and 201.43\% on the Bladder Tissue dataset, while maintaining real-time inference at 47.38 FPS per 1K matches, significantly outperforming other methods. 

To further evaluate matching robustness under low texture and large viewpoint changes, we select three representative image pairs from the zero-shot test set. Fig.~\ref{fig:exper} presents visual comparisons between our method and prior methods. Green lines indicate correct correspondences, while red lines denote initial matches rejected by RANSAC. The three rows illustrate challenges including large viewpoint shifts, low-texture regions, and significant camera advancement, respectively. Compared to other methods, LoFTR~\cite{sun2021loftr} and TransMatch~\cite{zhu2025transmatch} perform better but still suffer from sparse or incorrect matches in scenarios involving forward motion or textureless areas. In contrast, EndoMatcher consistently produces dense and accurate correspondences, robustly handling low textures and complex camera motion.

\subsubsection{Matching accuracy} % in downstream tasks

To evaluate the reliability of correspondences in downstream tasks, we assess matching accuracy on the Gastro-Matching dataset using Homography Estimation Accuracy (HEA) and Matching Direction Prediction Accuracy (MDPA), where MDPA reflects the ability to infer the correct relative motion direction from the predicted keypoint correspondences.
As shown in Table~\ref{tab:acc}, SIFT~\cite{lowe2004distinctive} + FLANN~\cite{muja2009flann} fails under endoscopic conditions, yielding extremely low HEA and MDPA scores. Sparse matchers such as SuperGlue~\cite{sarlin2020superglue} and LightGlue~\cite{lindenberger2023lightglue} achieve over 80\% in HEA@5px but remain below 60\% in MDPA, indicating limited robustness to motion variations. TransMatch~\cite{zhu2025transmatch}, specifically designed for gastroscopy, achieves the second highest accuracy in motion prediction, with 85.1\% HEA @ 3px and 76.0\% MDPA. In contrast, EndoMatcher achieves top performance across all metrics, improving MDPA by 9.40\% over TransMatch, and reliably delivers accurate correspondences without task-specific tuning.

\begin{table}[t]
\centering
\caption{Ablation on the Ours-Bronch and SCARED datasets using PCK (\%) at 5px, 10px, and 20px thresholds. The \textbf{best} and \underline{second-best} are highlighted.}
\label{tab:ablation}
\setlength{\tabcolsep}{3.7pt}
\renewcommand{\arraystretch}{1.5} 
\begin{tabular}{c|ccc|ccc}
\toprule
\multirow{2}{*}{} & \multicolumn{3}{c|}{\textbf{Ours-Bronch}} & \multicolumn{3}{c}{\textbf{SCARED}} \\
\cline{2-7}
 & @5px & @10px & @20px & @5px & @10px & @20px \\
\midrule
Ours w/o MS & 58.08 & 81.96 & \underline{93.18} & 97.50 & 98.64 & 99.20 \\
Ours w/o Clipping & 58.41 & 82.29 & 93.06 & 98.02 & \underline{99.17} & \underline{99.57} \\
Ours w/o DIB & \underline{58.65} & \underline{82.61} & 93.13 & \underline{98.03} & 99.14 & 99.56 \\
Ours & \textbf{59.26} & \textbf{83.46} & \textbf{94.32} & \textbf{98.27} & \textbf{99.30} & \textbf{99.65} \\
\bottomrule
\end{tabular}
\end{table}

\begin{table}[t!]
\centering
\caption{Zero-shot matching performance across different training configurations. The \textbf{best} and \underline{second-best} are highlighted.}
\label{tab:Ala_dataset}
\setlength{\tabcolsep}{3.3pt}
\renewcommand{\arraystretch}{1.6}
\begin{tabular}{c|ccc|ccc}
\toprule
\multirow{2}{*}{} & \multicolumn{3}{c|}{\textbf{Hamlyn Centre Dataset}} & \multicolumn{3}{c}{\textbf{Bladder Tissue Dataset}} \\
\cline{2-7}
& $N_{\text{pt}}$ & $N_{\text{inlier}}$ & $\text{KR}$ & $N_{\text{pt}}$ & $N_{\text{inlier}}$ & $\text{KR}$ \\
\midrule
Endo-Syn      &7987.15  &5848.57  &68.33  &7804.06 &5648.56	&\underline{67.99} \\
Endo-Real     &8012.48  &6047.60  &70.71  &7806.76	&5570.94	&66.21 \\
Endo-Mix6     &8095.39  &6117.87  &71.24  &8086.44	&5767.35	&66.34 \\
Endo-Mix6-MO  &\underline{8135.69}	&\underline{6190.75}  &\underline{72.17}  &\underline{8259.45}	&\underline{5951.85}	&67.40 \\
Endo-Mix6-PMO &\textbf{8190.21} &\textbf{6375.70} 	&\textbf{73.40}  &\textbf{8346.82} &\textbf{6116.20} 	&\textbf{68.75} \\

\bottomrule
\end{tabular}
\end{table}

\subsection{Ablation Studies}

\subsubsection{Robust multi-scale response loss}

We present quantitative results in Table~\ref{tab:ablation} using the Percentage of Correct Keypoints (PCK) to evaluate the effect of the proposed robust multi-scale response loss (RMSR) on keypoint localization performance and model robustness. Specifically, we assess two ablated variants of our network on the Ours-Bronch and SCARED datasets: \textbf{Ours w/o MS}, which removes multi-scale supervision and computes the loss solely at full resolution; and \textbf{Ours w/o Clipping}, which disables the residual clipping strategy and retains all residuals during training without excluding the lowest 20\%. Evaluations are conducted on manually curated validation and test sets. The results indicate that both components of RMSR, namely multi-scale supervision and residual clipping, contribute substantially to improved localization accuracy and robustness, particularly under large viewpoint variations and other challenging conditions. Specifically, multi-scale supervision enforces consistent correspondence learning across resolutions for better generalization under viewpoint changes, while residual clipping improves training focus and robustness by ignoring low-response outliers.

\subsubsection{Dual interaction block}
We evaluate the effectiveness of the proposed dual interaction block (DIB) in Table~\ref{tab:ablation}. Specifically, we compare two models: \textbf{Ours w/o DIB}, which retains all 12 transformer blocks from the original ViT, and the full model \textbf{Ours}, where the final 6 blocks are replaced by DIBs that alternate between self-attention and cross-attention to enhance inter-image feature interaction. Both training and evaluation are conducted on the Ours-Bronch and SCARED datasets. Results show that removing the cross-attention mechanism leads to a significant decrease in keypoint localization accuracy within the 20-pixel threshold. This result underscores the importance of cross-attention for modeling long-range dependencies and enabling contextual interaction between image pairs, which is particularly critical in endoscopic scenes with weak and repetitive textures.

\subsubsection{Training on diverse datasets}

We evaluate the impact of different training datasets and strategies on zero-shot generalization, as summarized in Table~\ref{tab:Ala_dataset}. Synthetic datasets (C3VD and EndoSLAM) offer large-scale, high-precision matching annotations, whereas real-world datasets (SCARED, EndoMapper, Colonoscopic, and Ours-Bronch) consist of narrow-lumen endoscopic videos characterized by rapid motion and lower annotation quality. To assess the effectiveness of our hybrid dataset design, we compare models trained under three configurations: (1) synthetic datasets only (\textbf{Endo-Syn}), (2) real-world datasets only (\textbf{Endo-Real}), and (3) the combined dataset (\textbf{Endo-Mix6}). Results show that integrating both real and synthetic data consistently enhances zero-shot matching performance compared to using either type alone. For example, on the Bladder dataset, Endo-Mix6 increases $N_{\text{inlier}}$ by 2.10\% and 3.53\% compared to Endo-Syn and Endo-Real, respectively. However, a straightforward combination of these data sources during training yields only marginal improvements, underscoring the importance of thoughtful dataset integration.

We further explore two advanced training strategies based on Endo-Mix6. First, a multi-objective optimization approach (\textbf{Endo-Mix6-MO}) balances heterogeneous data via Pareto-optimal~\cite{sener2018multi,ranftl2020towards} loss weighting. Second, a progressive multi-objective approach (\textbf{Endo-Mix6-PMO}), which first pretrains on synthetic data and then fine-tunes using multi-objective optimization on real-world data. Endo-Mix6-PMO achieves the best performance across both benchmark datasets. On the Hamlyn dataset, it improves KR by 5.07\% compared to Endo-Syn, and on the Bladder dataset, it increases $N_{\text{inlier}}$ by 9.79\% over Endo-Real.

\section{DISCUSSION AND CONCLUSION}

We present EndoMatcher, a generalizable endoscopic image matcher that uses a dual-branch, multi-scale Vision Transformer with dual interaction blocks for robust dense matching under challenging visual conditions. It is pretrained on our Endo-Mix6, a large-scale, multi-domain dataset containing 1.2M real and synthetic image pairs, and optimized via a progressive multi-objective training strategy. EndoMatcher achieves strong generalization without the need for task-specific fine-tuning. On zero-shot dense matching benchmarks, EndoMatcher achieves the highest match density, increasing the number of inlier matches ($N_{\text{inlier}}$) by 140.69\% and 201.43\%, respectively. On the Gastro-Matching dataset, EndoMatcher outperforms the domain-specific TransMatch by 9.40\% in Matching Direction Prediction Accuracy (MDPA), highlighting its geometric consistency and downstream reliability. In terms of efficiency, EndoMatcher supports real-time inference. All methods are tested on a single NVIDIA RTX 4090 GPU, where EndoMatcher reaches 47.38 FPS per 1K inlier matches for a 384×384 image pair, significantly outperforming LoFTR (12.83 FPS) and TransMatch, meeting the demands of real-time applications. Ablation studies further validate the contributions of key design components, including the multi-scale response loss, dual interaction blocks, and progressive training, all of which collectively enhance the model’s robustness, accuracy, and cross-domain generalization.

Despite its promising results, challenges remain for surgical deployment. Although inference is already efficient, further optimization is needed for latency-critical tasks such as intraoperative navigation and real-time SLAM. In future work, we plan to explore lightweight Transformer designs and model compression techniques to further reduce latency while preserving accuracy. Moreover, the Endo-Mix6 dataset has played a central role in bridging domain gaps and improving robustness. We will continue expanding the dataset to include more organs, imaging modalities, and pathological conditions.

% Overall, EndoMatcher sets a new SOTA for endoscopic feature matching, providing a robust, accurate, and efficient foundation for downstream robotic vision tasks in minimally invasive surgery.

\bibliographystyle{IEEEtran}
\bibliography{IEEEabrv,ref}

% \newpage

% \vspace{-10pt}
\begin{IEEEbiography}[{\includegraphics[width=1in,height=1.25in,clip,keepaspectratio]{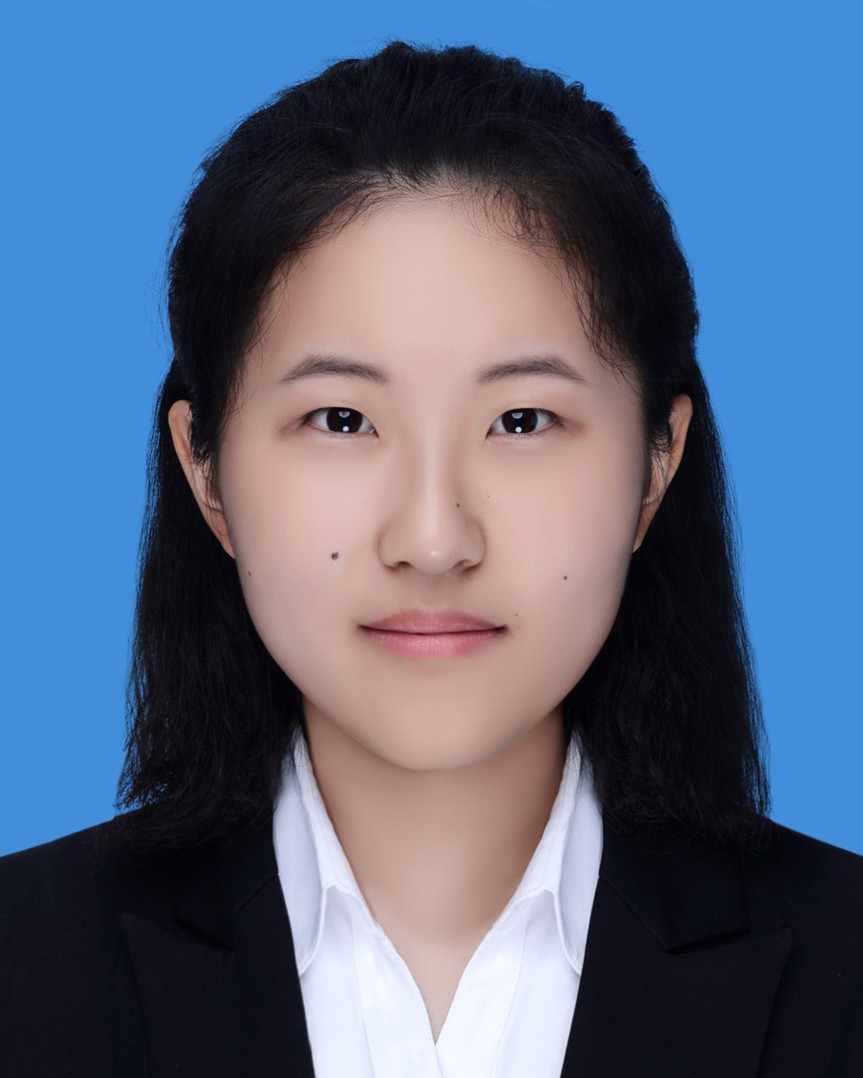}}]{Bingyu Yang}
received the B.Sc. degree in electronic information science and technology from Jilin University, Changchun, China, in 2022. She is currently pursuing the Ph.D. degree with the Institute of Automation, Chinese Academy of Sciences (CASIA), Beijing, China, under the supervision of Prof. Hongbin Liu. Her research focuses on endoscopy-CT fusion for surgical navigation in narrow luminal environments, encompassing deep learning-based CT image segmentation, endoscopic video point cloud reconstruction, and registration-based tracking.
\end{IEEEbiography}

\begin{IEEEbiography}[{\includegraphics[width=1in,height=1.25in,clip,keepaspectratio]{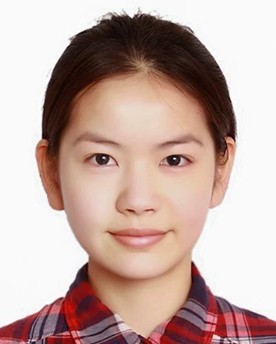}}]{Qingyao Tian}
received the B.Eng. degree in automation from North China Electric Power University, in 2021. She is currently pursuing the Ph.D. degree with the Institute of Automation, Chinese Academy of Sciences (CASIA), Beijing, China, under the supervision of Prof. Hongbin Liu. Her research focuses on vision-based surgical navigation, leveraging surgical scene foundation models and multimodal large models. 
\end{IEEEbiography}

\begin{IEEEbiography}[{\includegraphics[width=1in,height=1.25in,clip,keepaspectratio]{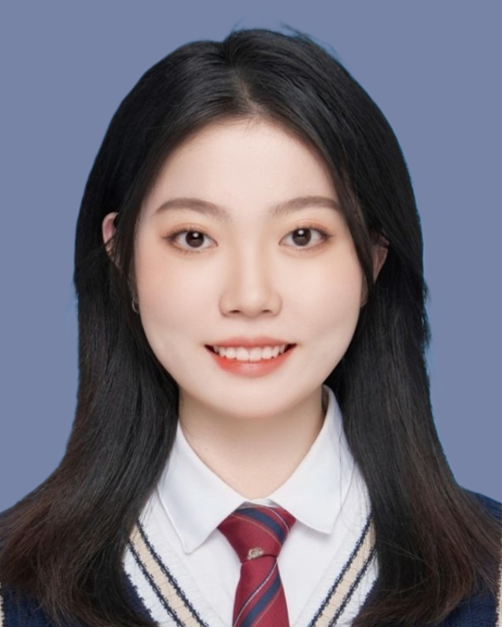}}]{Yimeng Geng}
received the B.Eng. degree in automation from Xi’an Jiaotong University, Xi’an, China, in 2022. She is currently pursuing the Ph.D. degree with the Institute of Automation, Chinese Academy of Sciences (CASIA), Beijing, China, under the supervision of Prof. Hongbin Liu. Her research focuses on multimodal fusion strategies for ultrasound image perception, encompassing deep learning-based image segmentation, deformable registration, and diagnostic recognition.
\end{IEEEbiography}

% \begin{IEEEbiography}[{\includegraphics[width=1in,height=1.25in,clip,keepaspectratio]{fig1}}]{Huai Liao}
% is a Professor and Chief Physician in the Department of Pulmonary and Critical Care Medicine at The First Affiliated Hospital of Sun Yat-sen University. His specialties include interventional pulmonology, lung cancer, and pleural diseases. He was a visiting scholar at Vanderbilt University and has authored over ten first-author publications in domestic and international journals.
% \end{IEEEbiography}

% \begin{IEEEbiography}[{\includegraphics[width=1in,height=1.25in,clip,keepaspectratio]{fig1}}]{Xinyan Huang}
% is a Physician in the Department of Pulmonary and Critical Care Medicine at The First Affiliated Hospital of Sun Yat-sen University. His clinical focus includes interventional pulmonology, lung cancer, and pleural diseases.
% \end{IEEEbiography}

\begin{IEEEbiographynophoto}{Huai Liao}
is a Professor and Chief Physician in the Department of Pulmonary and Critical Care Medicine at The First Affiliated Hospital of Sun Yat-sen University. His specialties include interventional pulmonology, lung cancer, and pleural diseases. He was a visiting scholar at Vanderbilt University and has authored over ten first-author publications in domestic and international journals.
\end{IEEEbiographynophoto}

\begin{IEEEbiographynophoto}{Xinyan Huang}
is a Physician in the Department of Pulmonary and Critical Care Medicine at The First Affiliated Hospital of Sun Yat-sen University. His clinical focus includes interventional pulmonology, lung cancer, and pleural diseases.
\end{IEEEbiographynophoto}

\begin{IEEEbiography}[{\includegraphics[width=1in,height=1.25in,clip,keepaspectratio]{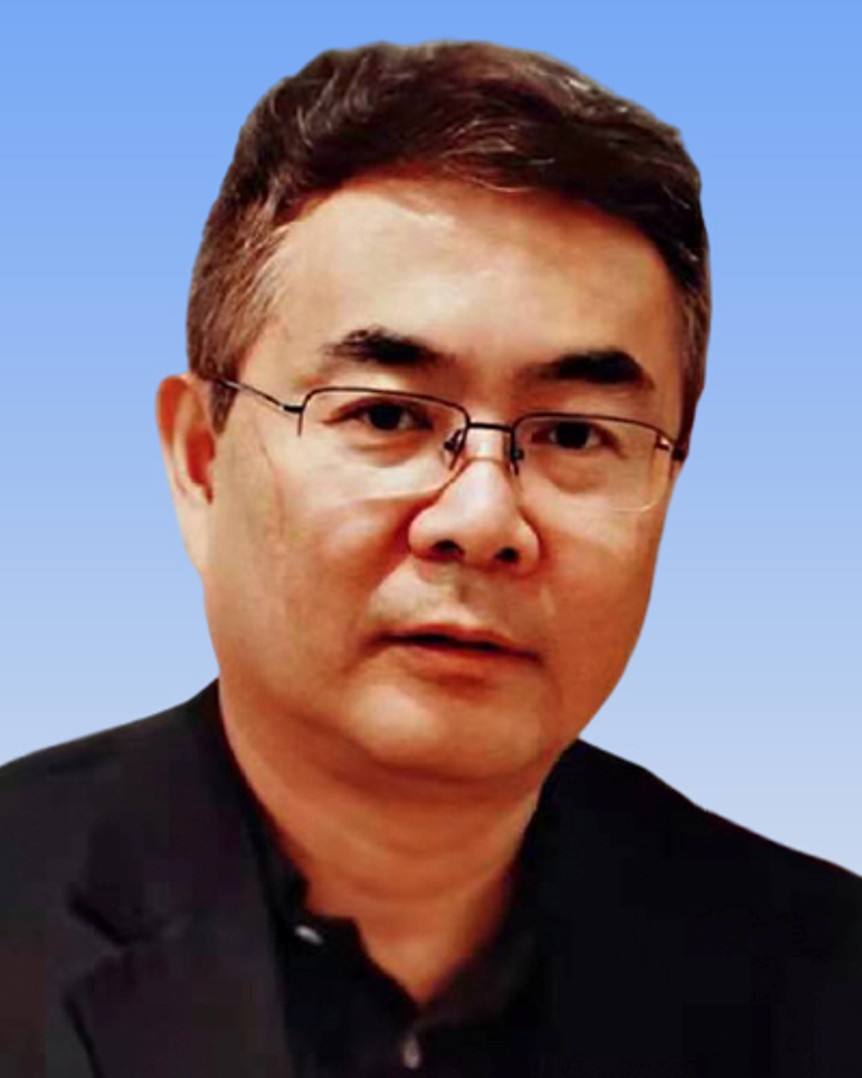}}]{Jiebo Luo}
(Fellow, IEEE) is the Albert Arendt Hopeman Professor of Engineering and Professor of Computer Science at the University of Rochester, Rochester, NY, USA, which he joined after a prolific career of fifteen years with Kodak Research Laboratories in 2011. He has authored over 600 technical papers and holds more than 90 U.S. patents. His research interests include computer vision, NLP, machine learning, data mining, multimedia, computational social science, and digital health. Prof. Luo has been involved in numerous technical conferences, including serving as Program Co-Chair of ACM Multimedia 2010, IEEE CVPR 2012, ACM ICMR 2016, and IEEE ICIP 2017, and General Co-Chair of ACM Multimedia 2018, and IEEE ICME 2024. He served on the editorial boards of the IEEE Transactions on Pattern Analysis and Machine Intelligence, IEEE Transactions on Multimedia, IEEE Transactions on Circuits and Systems for Video Technology, IEEE Transactions on Big Data, ACM Transactions on Intelligent Systems and Technology, Pattern Recognition, and Intelligent Medicine. He was the Editor-in-Chief of IEEE Transactions on Multimedia from 2020 to 2022. Professor Luo is also a Fellow of ACM, AAAI, AIMBE, SPIE, and IAPR, and a Member of Academia Europaea and the US National Academy of Inventors.
\end{IEEEbiography}

\begin{IEEEbiography}[{\includegraphics[width=1in,height=1.25in,clip,keepaspectratio]{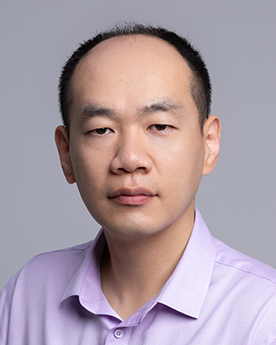}}]{Hongbin Liu}
is a Professor with the Institute of Automation (IA), Chinese Academy of Sciences (CAS), Beijing, China, and the Director of the Centre for Artificial Intelligence and Robotics (CAIR), Hong Kong Institute of Science and Innovation, Hong Kong SAR. Dr. Liu is also an Adjunct Reader and the Director of the Haptic Mechatronics and Medical Robotics (HaMMeR) Laboratory, School of Biomedical Engineering and Imaging Sciences, King’s College London (KCL), London, U.K. His group has been focusing on research and development of medical robotic systems with advanced haptic perception and interaction capabilities, to enable safer and more effective minimally invasive diagnosis and treatment for patients. His research has led to the clinical translation of a series of flexible robotic endoscopic systems for applications such as colonoscopy, bronchoscopy, and vascular surgeries.
\end{IEEEbiography}

\vfill

\end{document}